\documentclass{article} 
\usepackage{iclr2027_conference,times}

\usepackage{amsmath,amsfonts,bm}

\def\eqref#1{equation~\ref{#1}}

\def\1{\bm{1}}

\DeclareMathAlphabet{\mathsfit}{\encodingdefault}{\sfdefault}{m}{sl}
\SetMathAlphabet{\mathsfit}{bold}{\encodingdefault}{\sfdefault}{bx}{n}

\newcommand{\R}{\mathbb{R}}

\usepackage{hyperref}
\usepackage{url}
\usepackage{graphicx}
\usepackage{subcaption}
\usepackage{booktabs, multirow, amsmath}

\title{MaskCoFT: Masked Co-Adaptive Fine-Tuning for Memory-Efficient MoE Inference}

\author{
Junfeng Wu\thanks{Department of Industrial and Systems Engineering,
Rensselaer Polytechnic Institute.}
\And
Zehao Fan\thanks{Department of Electrical, Computer, and Systems
Engineering, Rensselaer Polytechnic Institute}
\And
Hadjer Benmeziane\thanks{IBM Research}
\And
Kaoutar El Maghraoui\footnotemark[3]
\And
Liu Liu\footnotemark[2]
\And
Yinan Wang\footnotemark[1]
}

\begin{document}

\maketitle

\begin{abstract}

Mixture-of-experts (MoE) language models often exceed the memory of a single GPU. Expert offloading keeps most experts in host memory and loads them on demand, so decoding speed depends on how many experts each token must fetch. Caching and prefetching reduce this cost only as far as the routing allows. Router-only fine-tuning can reshape the routing to reuse experts, but it keeps the experts frozen, so they cannot adapt to the tokens the new routing sends them. We propose \textbf{MaskCoFT}, a masked co-adaptive fine-tuning method that trains routers and experts together with the cross-entropy loss alone. During fine-tuning, a learnable binary mask restricts the Top-$K$ routing of each layer to a subset of experts, and the experts adapt to the tokens redirected to them. At inference, the learned mask becomes a soft prior that re-ranks experts, so every expert remains selectable. We simulate a GPU cache of 4 experts per layer for Mixtral-8×7B and 12 for DeepSeek-V2-Lite. MaskCoFT cuts expert fetches per token by 23.7\% and 10.1\% relative to the base model. In real offloading system serving, it lowers the time per output token by up to 16.4\% and 5.5\%, respectively. Its average accuracy over nine benchmarks stays above the base model by 0.92 and 0.53 points.
\end{abstract}

\section{Introduction}
\label{introduction}

\begin{figure}[htbp]
\centering
\includegraphics[width=\linewidth]{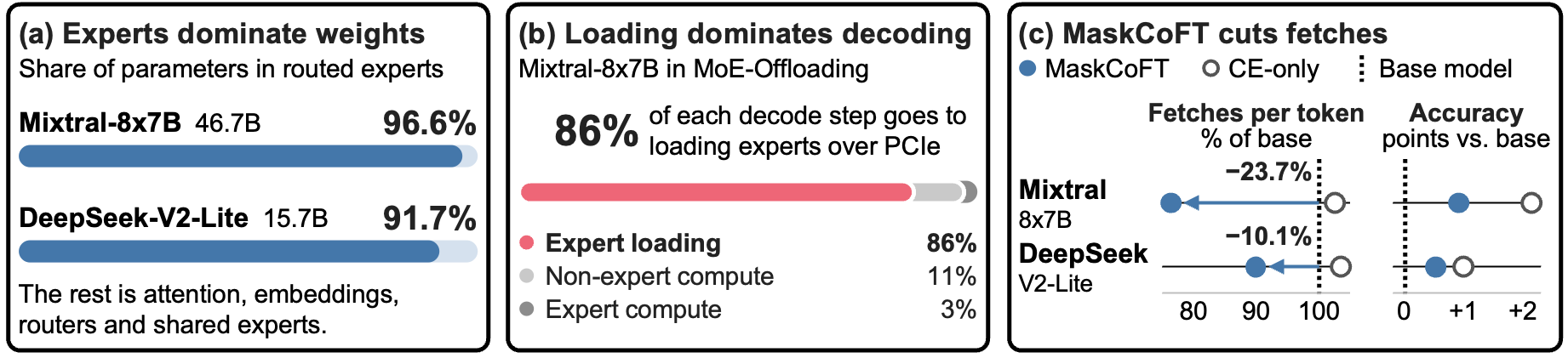}
\caption{\textbf{Routed expert weights dominate memory, and loading them dominates offloaded decoding latency. MaskCoFT reduces expert fetches while preserving accuracy.} (a) Routed experts hold 96.6\% of Mixtral-8×7B's parameters and 91.7\% of DeepSeek-V2-Lite's. (b) Decode latency breakdown of Mixtral-8×7B in MoE-Offloading~\citep{eliseev2023fast}, measured with NVIDIA Nsight Systems~\citep{nvidia_nsight}. (c) Expert fetches per token and average accuracy over nine benchmarks under GPU cache holding 4 experts per layer for Mixtral-8×7B and 12 for DeepSeek-V2-Lite, both relative to the base model. The GPU cache holds 4 experts per layer for Mixtral-8×7B and 12 for DeepSeek-V2-Lite. MaskCoFT cuts fetches by 23.7\% on Mixtral-8×7B and by 10.1\% on DeepSeek-V2-Lite. It keeps average accuracy above the base model.}
    \label{fig:moe_overview}
\end{figure}

Mixture-of-experts (MoE) architectures are now widely used to scale large language models (LLMs). They expand model capacity while activating only a small subset of experts per token~\citep{fedus2022switchtransformersscalingtrillion,jiang2024mixtral}. Although sparse activation reduces computation per token, memory still scales with the total number of experts~\citep{skliar2025mixture}. Routed experts hold 96.6\% of the parameters in Mixtral-8×7B~\citep{jiang2024mixtral} and 91.7\% in DeepSeek-V2-Lite~\citep{liu2024deepseek} (Figure~\ref{fig:moe_overview}a). When these weights exceed GPU memory, MoE offloading systems cache a subset of experts in GPU memory and keep the rest in host memory~\citep{xue2024moe,kamahori2025fiddler}.

During autoregressive decoding, expert selections can vary across successive tokens. Selected experts absent from the GPU cache must be loaded from host memory before execution, with cached experts evicted as needed to free space. These repeated expert migrations introduce data transfer latency, limiting decoding efficiency. In MoE-Offloading~\citep{eliseev2023fast}, expert loading over PCIe takes 86\% of the decoding time of Mixtral-8×7B, while expert computation takes only 3\% (Figure~\ref{fig:moe_overview}b). Prior offloading systems report the same bottleneck~\citep{hwang2024pre,kamahori2025fiddler,xue2024moe}. On memory-constrained devices, decoding speed therefore depends mainly on the number of expert migrations.

To reduce expert-loading latency, existing systems employ expert caching, predictive prefetching, and memory-aware expert scheduling~\citep{hwang2024pre,kong2024swapmoe,he2024expertflow}. Heterogeneous computing systems further reduce weight movement by executing selected experts near memory and transferring activations instead of expert weights~\citep{kim2024monde,fan2025contextawaremixtureofexpertsinferencecxlenabled}. However, the effectiveness of caching and prefetching remains constrained by the model's expert access pattern, GPU memory capacity, and transfer bandwidth. When decoding requests exhibit limited expert reuse, a small GPU cache can still incur frequent expert misses. Even accurate prefetching cannot fully hide their cost when the required transfers take longer than the computation available for overlap. These limitations motivate a complementary approach: fine-tuning the model to concentrate tokens on fewer experts, which increases expert reuse under a fixed GPU memory budget. However, those experts must then process tokens they were not trained on, so accuracy may drop. We therefore fine-tune routers and experts jointly under binary routing masks. The masks steer tokens toward a subset of experts, and those experts adapt to the redirected tokens.


The closest prior work, ReMoE~\citep{zhu2026remoe}, fine-tunes only the routers and keeps the experts frozen. It adds a temporal locality regularizer, which rewards reusing the experts of recent tokens, and a semantic anchor loss, which keeps routing close to the pretrained router.  This design leaves two questions open. First, its reuse comes from nearby tokens selecting the same experts, so its benefit may shrink when adjacent tokens need different experts. Second, the frozen experts cannot adapt to the tokens that the updated router sends them, which may limit how far routing can concentrate before accuracy drops.

We propose \textbf{MaskCoFT} (Masked Co-adaptive Fine-Tuning) to address both questions. During fine-tuning, a learnable binary mask restricts the top-\(K\) routing of each layer to a sampled, fixed-size concentration set of experts that all tokens share. Because all tokens share the set, reuse no longer depends on nearby tokens selecting the same experts. We fine-tune the experts jointly with the routers, so the experts adapt to the tokens redirected into the set. At inference, we lift the hard restriction and use the learned mask as a prior to re-rank experts. Routing then favors the concentration set but can still be able to select any expert.

The proposed MaskCoFT is evaluated on two MoE architectures with different expert granularities: the coarse-grained Mixtral-8×7B and the fine-grained DeepSeek-V2-Lite. Coarse-grained MoEs use fewer, larger experts, whereas fine-grained MoEs use more, smaller experts to encourage finer specialization~\citep{liang2026not, liu2024deepseek, wang2024let}. Our experiments assess whether router-expert co-adaptation can reduce expert migrations while preserving model quality across both granularities.

Our contributions are summarized as follows:

(1) We propose MaskCoFT, which fine-tunes routers and experts jointly under a learnable binary routing mask. The mask restricts each layer to a shared set of experts during fine-tuning and becomes a soft prior that re-ranks experts at inference.

(2) MaskCoFT trains with the cross-entropy loss alone. It needs no auxiliary locality or load-balancing loss, so there are no loss hyperparameters to tune.

(3) We evaluate MaskCoFT on the coarse-grained Mixtral-8×7B and the fine-grained DeepSeek-V2-Lite. With an LFU expert cache, it cuts expert fetches per token by 23.7\% and 10.1\%, respectively, relative to the base model. Its fetch reductions hold under LRU, LFU and FIFO at every cache budget we test. In MoE-Offloading, it lowers the time per output token by up to 16.4\% and 5.5\%. Its average accuracy over nine benchmarks stays above the base model on both.

\section{Related Work}
\label{related work}


\textbf{Offloading systems for MoE inference.}
MoE offloading systems reduce the cost of expert migration through caching, prefetching and heterogeneous execution. MoE-Offloading~\citep{eliseev2023fast}, MoE-Infinity~\citep{xue2024moe}, Pre-gated MoE~\citep{hwang2024pre}, SwapMoE~\citep{kong2024swapmoe}, ExpertFlow~\citep{he2024expertflow} and FineMoE~\citep{yu2026taming} improve cache replacement and predictive prefetching. Fiddler~\citep{kamahori2025fiddler}, DAOP~\citep{zhang2025daop} and MoNDE~\citep{kim2024monde} run less frequently used experts on the CPU or on near-data processing (NDP) units. Cache-aware routing favors experts that are already in the cache at inference, without training~\citep{skliar2025mixture}. MaskCoFT instead changes the routing through fine-tuning and can be compatible with different cache policies.

\textbf{Training methods for efficient MoE inference.}
Other work trains MoE models to be cheaper to serve under memory limits by shaping routing and expert usage. Oracle-MoE~\citep{zhou2025oracle} routes tokens by their clusters in semantic space, so that consecutive tokens tend to activate the same experts. StickyMoE~\citep{kayyam2026sticky} adds a routing-consistency regularizer to the cross-entropy and load-balancing losses, so routers and experts co-adapt during pretraining. Both methods train an MoE model from scratch, and their experiments use small GPT-style models with up to only 2B parameters. ReMoE~\citep{zhu2026remoe} fine-tunes only the routers of a pretrained MoE and keeps the experts frozen. Its loss adds a temporal locality regularizer and a semantic anchor, a KL penalty that keeps routing close to the pretrained router. Concurrent work~\citep{wu2026cache} fine-tunes auxiliary cache routers jointly with the MoE backbone and keeps the native top-$K$ selection. It learns which experts to keep in the cache. MaskCoFT instead changes which experts the tokens select to make them concentrate. It fine-tunes the routers and experts of a pretrained MoE with the cross-entropy loss alone, and a learned mask concentrates token activations on a subset of experts in each layer.

\textbf{Expert pruning and learnable masking.}
Prior work employs pruning and masking to facilitate network sparsity and computational efficiency while preserving model capacity. MaskConnect~\citep{ahmed2018maskconnect} jointly learns binary connectivity and network weights for convolutional neural networks to search a sparse architecture. For MoE models, \citet{lu2024not} reduce memory and computational cost of model deployment by pruning and skipping experts. BEAM~\citep{wu2026beam} reduces per-token computation through a learnable token-adaptive mask and achieves a lower number of activated experts for each token on average. These methods primarily reduce model size and computational redundancy rather than expert migration latency in memory-constrained MoE offloading systems. Fewer active experts per token do not necessarily imply greater reuse across tokens. Our work uses binary routing masks during joint fine-tuning of experts and routers to encourage concentrated expert activations, targeting the reduction of expert transfers during offloaded decoding.

\section{Methodology}
\label{headings}

\subsection{Preliminaries}
\textbf{Mixture-of-experts layer.}
An MoE model has $L$ MoE layers. Each MoE layer replaces the feed-forward block of a Transformer layer with $E$ routed experts $\mathcal{F}_1,\dots,\mathcal{F}_E$ and a router. The router has weight $\mathbf{W}\in\R^{E\times d_h}$, where $d_h$ is the hidden dimension. For the hidden state $\mathbf{x}\in\R^{d_h}$ of one token, the router computes the routing probabilities $\mathbf{p}=\operatorname{softmax}(\mathbf{W}\mathbf{x})\in\R^{E}$. Top-$K$ routing activates the $K$ experts with the largest probabilities and sums their gated outputs:
\begin{equation}
\mathcal{A}=\operatorname{Top\text{-}K}(\mathbf{p}),\qquad
\mathbf{y}=\sum_{i\in\mathcal{A}} g_i\,\mathcal{F}_i(\mathbf{x}),
\label{eq:topk}
\end{equation}
where $\mathcal{A}$ is the set of active experts. DeepSeek-V2-Lite uses the gate weight $g_i=p_i$. Mixtral-8×7B renormalizes it over the active set, $g_i=p_i/\sum_{j\in\mathcal{A}}p_j$. DeepSeek-V2-Lite also has shared experts per layer, which process every token and add their outputs to $\mathbf{y}$. Routing and masking involve only the routed experts. Each layer has its own router and, in MaskCoFT, its own mask. We drop the layer index $\ell$ when the layer is clear from context.

Top-$K$ routing depends on each token, so successive tokens can select different experts. The selected experts are then often missing from a small GPU cache. In the offloading systems we study, every miss triggers a fetch over PCIe.



\subsection{MaskCoFT: Masked Co-adaptive Fine-tuning}

\begin{figure}[htbp]
\centering
\includegraphics[width=\linewidth]{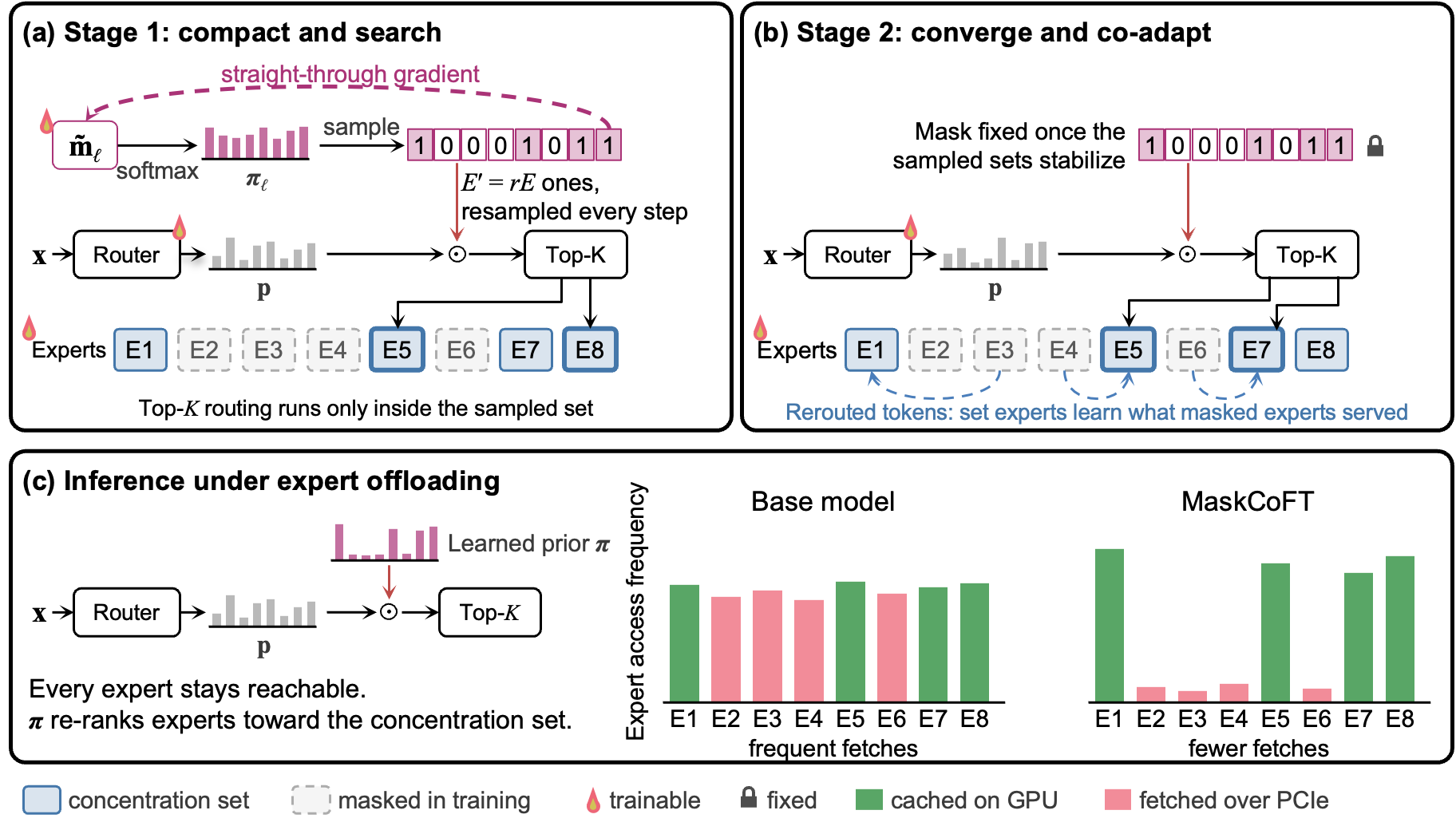}
\caption{\textbf{MaskCoFT} learns a concentration set of experts per layer and co-adapts routers and experts. (a)~At every step, Stage~1 samples a set of $E'$ experts from $\boldsymbol{\pi}_\ell$ and runs Top-$K$ routing only inside it. A straight-through gradient, accumulated over $T$ steps, updates $\tilde{\mathbf{m}}_\ell$. (b)~Stage~2 fixes the sets of all layers, then the routers and experts adapt to the rerouted tokens. (c)~At inference, the learned prior $\boldsymbol{\pi}$ re-ranks experts toward the concentration set, so decoding fetches fewer experts.}
\label{fig:flowchart}
\end{figure}

MaskCoFT fine-tunes the routers and experts in two stages (Figure~\ref{fig:flowchart}a, b) and re-ranks experts at inference (Figure~\ref{fig:flowchart}c). Both stages restrict routing in each layer to a \emph{concentration set}, a subset of $E'$ experts chosen by a learned mask. Stage~1 searches for these sets. Stage~2 fixes them and lets the routers and experts co-adapt.


\textbf{Stage 1: Compact and search.}
For each MoE layer $\ell$, we keep a learnable real-valued mask $\tilde{\mathbf{m}}_\ell\in\R^{E}$ with equal entries at initialization. A softmax turns $\tilde{\mathbf{m}}_\ell$ into a sampling distribution $\boldsymbol{\pi}_\ell$, so $\pi_{\ell,i}=1/E$ at the start. We draw $E'=rE$ distinct experts from $\boldsymbol{\pi}_\ell$ by multinomial sampling without replacement. The compact ratio $r$ is the fraction of routed experts that a concentration set keeps. We choose $r$ so that $K\le E'\le E$. The drawn experts define a binary mask $\mathbf{m}_\ell\in\{0,1\}^{E}$ with exactly $E'$ ones. Top-$K$ routing then runs only inside this concentration set:
\begin{equation}
\boldsymbol{\pi}=\operatorname{softmax}(\tilde{\mathbf{m}}),\qquad
\mathbf{m}=\operatorname{MultinomialSample}(\boldsymbol{\pi},E'),\qquad
\mathcal{A}=\operatorname{Top\text{-}K}(\mathbf{p}\odot\mathbf{m}).
\label{eq:stage1}
\end{equation}
The output follows Eq.~(\ref{eq:topk}) with gate weights computed from $\mathbf{p}\odot\mathbf{m}$. On $\mathcal{A}$, these values equal the gate weights computed from $\mathbf{p}$, since $m_i=1$ for every $i\in\mathcal{A}$. The product matters in the backward pass, where it gives the loss a gradient with respect to $\mathbf{m}$ (see \textbf{Mask optimization} below). All tokens in a training batch share the same mask. We resample the binary masks at every step, so Stage~1 explores different concentration sets. The routers and experts update at every step. For the real-valued masks, we accumulate the gradient over $T$ steps and then update them once, so each update averages over $T$ sampled sets. Refer to Appendix \ref{Appendix_C} for the convergence of concentration expert sets.

\textbf{Stage 2: Converge and co-adapt.}
Once the sampled sets stabilize after Stage 1, we fix each layer's binary mask and real-valued mask to its current concentration set. We stop updating $\tilde{\mathbf{m}}$ and continue to fine-tune the routers and experts. The routers learn to split tokens among the experts of the fixed set. Tokens that masked experts used to serve now reach experts inside the set. The low-rank adapters (LoRA)~\citep{hu2021lora} of these experts learn to process them from the fine-tuning data (Figure~\ref{fig:flowchart}b). We call this knowledge transfer to the concentration set. Router-only methods such as ReMoE~\citep{zhu2026remoe} keep the experts frozen, so their experts cannot adapt to rerouted tokens. Stage~2 aims to recover accuracy under the restricted routing. Refer to Section \ref{Experimental Setup} and Appendix~\ref{Appendix_B}  for the detailed training setup.


\textbf{Training objective.}
Both stages minimize the standard cross-entropy loss $\mathcal{L}$ without an auxiliary load-balancing loss~\citep{lepikhin2020gshard}. A load-balancing loss spreads tokens evenly across experts. Even spreading works against our goal of concentrating expert activations to reduce fetches during memory-constrained decoding~\citep{zhu2026remoe}.

\textbf{Mask optimization.}
Sampling the binary mask is not differentiable. We therefore use a straight-through estimator (STE), which treats the sampling step as the identity in the backward pass~\citep{ahmed2018maskconnect,bengio2013estimating}. The loss $\mathcal{L}$ reaches $\mathbf{m}$ through the gate weights of the active experts. Because all tokens share the mask, $\partial\mathcal{L}/\partial\mathbf{m}$ sums the contributions of every token in the batch. The STE passes this gradient to $\boldsymbol{\pi}$, and the softmax Jacobian carries it to $\tilde{\mathbf{m}}$:
\begin{equation}
\frac{\partial\mathcal{L}}{\partial\boldsymbol{\pi}}\approx\frac{\partial\mathcal{L}}{\partial\mathbf{m}},\qquad
\frac{\partial\mathcal{L}}{\partial\tilde{\mathbf{m}}}\approx\bigl(\operatorname{diag}(\boldsymbol{\pi})-\boldsymbol{\pi}\boldsymbol{\pi}^{\top}\bigr)\frac{\partial\mathcal{L}}{\partial\mathbf{m}}.
\label{eq:ste}
\end{equation}



\textbf{Evaluation and inference.}
At evaluation and inference, we replace the binary mask with the learned sampling distribution $\boldsymbol{\pi}$, which now acts as a prior that re-ranks experts (Figure~\ref{fig:flowchart}c):
\begin{equation}
\mathcal{A}=\operatorname{Top\text{-}K}(\mathbf{p}\odot\boldsymbol{\pi}).
\label{eq:inference}
\end{equation}

Every $\pi_i$ is positive, so the prior excludes no expert by construction. Top-$K$ still selects an expert as active outside the concentration set when its product $p_i\pi_i$ ranks among the $K$ largest. The active experts keep the gate weights $g_i$ of Eq.~(\ref{eq:topk}), computed from $\mathbf{p}$. The prior changes which experts are active but not their gate weights. A hard mask would force every token into the set. The soft prior lets a token use an outside expert when the router strongly prefers it. 



\section{Experiment}
\label{evaluation}
We evaluate our proposed method in the following aspects: (1) a stack of nine benchmarks for model capacity evaluation; (2) offloading system simulation based on the expert selection trace; (3) latency breakdown and end-to-end performance measurement on a real MoE offloading system.     

\subsection{Experimental Setup}
\label{Experimental Setup}
\textbf{Models.} We evaluate MaskCoFT on two MoE models that differ in size and expert granularity: Mixtral-8×7B~\citep{jiang2024mixtral} and DeepSeek-V2-Lite~\citep{liu2024deepseek}. Appendix~\ref{Appendix_A} gives their details.

\textbf{Dataset.} We fine-tune on OpenHermes-2.5~\citep{teknium2023openhermes}, a multi-turn instruction and chat corpus that covers general chat, reasoning, code and math. We follow the preprocessing of ReMoE~\citep{zhu2026remoe} and sample 100K examples for training and 1K for held-out validation.

\textbf{Fine-tuning.} We fine-tune all routed experts with LoRA adapters~\citep{hu2021lora}, as well as the shared experts of DeepSeek-V2-Lite. We fully fine-tune the router of each MoE layer. Both models use the compact ratio $r=75\%$, the AdamW optimizer~\citep{loshchilov2017decoupled} and BF16 precision. We train on 2 H200 GPUs with PyTorch Distributed Data Parallel (DDP)~\citep{li2020pytorch} and a batch size of 32 per device. Appendix~\ref{Appendix_B} lists the other hyperparameters.

\textbf{Compared methods.} We compare MaskCoFT with the pretrained base model (Baseline), fine-tuning with the cross-entropy loss only (CE-only) and the results reported by ReMoE~\citep{zhu2026remoe}. CE-only differs from MaskCoFT only in the mask.

\textbf{Capacity benchmark.}
We measure accuracy with the EleutherAI LM Evaluation Harness~\citep{eval-harness} on nine benchmarks. Seven are 0-shot multiple-choice tasks on language understanding and reasoning: MMLU, ARC-E, ARC-C, HellaSwag, WinoGrande, PIQA and BoolQ. GSM8K tests math reasoning with 5 shots, and HumanEval tests code generation with 0 shots. We report the flexible-extract score for GSM8K and pass@1 for HumanEval, as ReMoE does~\citep{zhu2026remoe}. Flexible-extract takes the last number in the output as the answer. Strict-match, which Table~\ref{tab:compare_remoe} also reports, accepts only the number that follows the ``\#\#\#\#'' marker used in the few-shot examples.

\textbf{Simulation.}
We measure how well each model's traffic fits a per-layer expert cache with a trace-driven simulator. Following the experiment and metric definitions of ReMoE~\citep{zhu2026remoe}, we sample 128 sequences from the ShareGPT V3 unfiltered dataset~\citep{anon2023sharegpt}. All models receive the same prompts and use greedy decoding with batch size 1 and at most 64 new tokens. We record the selected experts at each decode step. 
Each layer keeps a cache of $B$ routed experts, with $K\le B<E$. Shared experts stay on the GPU and do not count toward $B$. Caches start empty and persist across prompts, and initial fills count as misses. Each miss triggers one expert fetch and, if the cache is full, an eviction under the least-recently-used (LRU), least-frequently-used (LFU) or first-in-first-out (FIFO) policy. Over $N$ decode steps, we report the hit rate (HR) and the expert fetches per token (Fetches/tok):
\begin{equation}
\mathrm{HR}=\frac{1}{NLK}\sum_{\ell=1}^{L}\sum_{t=1}^{N} h_{\ell,t},\qquad
\mathrm{Fetches/tok}=\frac{1}{N}\sum_{\ell=1}^{L}\sum_{t=1}^{N} f_{\ell,t},
\label{eq:metrics}
\end{equation}
where $h_{\ell,t}$ counts the active experts at layer $\ell$ and step $t$ that are already in the cache, and $f_{\ell,t}=K-h_{\ell,t}$ counts the misses. The two metrics therefore satisfy $\mathrm{Fetches/tok}=LK(1-\mathrm{HR})$.


\textbf{System measurement.}
We measure the models served in MoE-Offloading~\citep{eliseev2023fast} with the NVIDIA Nsight system profiler~\citep{nvidia_nsight}. We report the system-level metrics, including Time-to-First-Token (TTFT), Time-per-Output-Token (TPOT), and end-to-end Throughput. For each model, we select 32 prompts that contain at least 128 tokens from
ShareGPT V3 unfiltered dataset, and
truncate each prompt to the first 128 tokens to fix the input token length. We evaluate output lengths of 64 and 128 tokens and report the mean measurement across requests.      

\subsection{Model Capacity Evaluation}
\begin{table}[htbp]
  \centering
  \caption{Accuracy (\%) on nine benchmarks. GSM8K is 5-shot with flexible-extract scoring. HumanEval is 0-shot pass@1. The other seven tasks are 0-shot multiple choice. Avg.\ is the mean over the nine benchmarks. Bold marks the best of the three models in each column. The $\Delta$ rows give MaskCoFT minus each reference, in percentage points.}
  \label{tab:capacity}
  \small
  \setlength{\tabcolsep}{3.5pt}
  \resizebox{\textwidth}{!}{%
  \begin{tabular}{@{}lcccccccccc@{}}
    \toprule
    & \multicolumn{2}{c}{Generative} & \multicolumn{7}{c}{Multiple choice} & \\
    \cmidrule(lr){2-3}\cmidrule(lr){4-10}
    Method & GSM8K & HumanEval & MMLU & ARC-E & ARC-C & HellaSwag & WinoGrande & PIQA & BoolQ & Avg. \\
    \midrule
    \multicolumn{11}{@{}l}{\textit{\textbf{DeepSeek-V2-Lite}}} \\
    Baseline & \textbf{38.36} & 26.83 & \textbf{55.53} & 74.37 & 45.99 & 77.73 & 71.27 & 80.25 & 80.34 & 61.19 \\
    CE-only & 37.38 & \textbf{28.05} & 54.73 & \textbf{77.69} & \textbf{50.51} & \textbf{79.07} & \textbf{72.22} & \textbf{80.69} & 79.30 & \textbf{62.18} \\
    MaskCoFT (Ours) & 38.06 & \textbf{28.05} & 53.36 & 75.72 & 49.57 & 76.06 & 71.82 & 80.36 & \textbf{82.45} & 61.72 \\
    $\Delta_{\mathrm{Baseline}}$ & $-0.30$ & $+1.22$ & $-2.17$ & $+1.35$ & $+3.58$ & $-1.67$ & $+0.55$ & $+0.11$ & $+2.11$ & $+0.53$ \\
    $\Delta_{\text{CE-only}}$ & $+0.68$ & $0.00$ & $-1.37$ & $-1.97$ & $-0.94$ & $-3.01$ & $-0.40$ & $-0.33$ & $+3.15$ & $-0.46$ \\
    \midrule
    \multicolumn{11}{@{}l}{\textit{\textbf{Mixtral-8×7B}}} \\
    Baseline & 58.68 & 35.37 & 68.16 & 82.53 & 59.47 & 84.24 & \textbf{77.11} & 83.30 & 85.96 & 70.54 \\
    CE-only & \textbf{68.46} & \textbf{38.29} & \textbf{68.77} & \textbf{84.18} & \textbf{61.95} & \textbf{84.30} & 76.48 & 83.68 & 87.98 & \textbf{72.68} \\
    MaskCoFT (Ours) & 66.19 & 36.10 & 67.18 & 82.91 & 58.45 & 83.56 & 76.56 & \textbf{83.73} & \textbf{88.44} & 71.46 \\
    \addlinespace[2pt]
    $\Delta_{\mathrm{Baseline}}$ & $+7.51$ & $+0.73$ & $-0.98$ & $+0.38$ & $-1.02$ & $-0.68$ & $-0.55$ & $+0.43$ & $+2.48$ & $+0.92$ \\
    $\Delta_{\text{CE-only}}$ & $-2.27$ & $-2.19$ & $-1.59$ & $-1.27$ & $-3.50$ & $-0.74$ & $+0.08$ & $+0.05$ & $+0.46$ & $-1.22$ \\
    \bottomrule
  \end{tabular}}
\end{table}
Table~\ref{tab:capacity} reports accuracy on nine benchmarks. Averaged over them, MaskCoFT beats the base model by 0.53 points on DeepSeek-V2-Lite and by 0.92 points on Mixtral-8×7B. It trails CE-only by 0.46 and 1.22 points, respectively. The $\Delta_{\mathrm{Baseline}}$ rows report the difference between MaskCoFT and the base model. For both architectures, MaskCoFT achieves performance generally better than the base model and slightly worse than the CE-only fine-tuned model. CE-only outperforms the base model on most benchmarks, suggesting that fine-tuning on the sampled dataset can improve downstream performance even without a load-balancing loss. To isolate the effect of masking from the benefits of fine-tuning, we compare MaskCoFT with CE-only and report the differences in $\Delta_{\text{CE-only}}$. Both methods use the same cross-entropy objective and differ only in the use of masking. The results suggest that MaskCoFT largely preserves the performance benefits of fine-tuning.

The results also reveal differences between the two MoE structures of different granularity. On the generative benchmarks GSM8K and HumanEval, MaskCoFT scores about 2.2 points below CE-only on the coarse-grained Mixtral-8×7B. However, the fine-grained DeepSeek-V2-Lite matches or exceeds CE-only on both. During fine-tuning at the same compact ratio of 75$\%$, the binary mask excludes 2 of the 8 large experts per layer of Mixtral-8×7B. On the other hand, it excludes 16 of the 64 small experts per layer of DeepSeek-V2-Lite, whose two shared experts also stay active for every token per layer. \citet{skliar2025mixture} report that MoEs with more, smaller active experts are more resilient to changes in expert selection. 


\subsection{Simulation Results}
\begin{table}[htbp]
  \centering
  \caption{Trace-driven cache simulation. HR is the hit rate, and Fetches/tok is the number of expert fetches per token (Eq.~(\ref{eq:metrics})). The GPU cache holds $B=12$ experts per layer for DeepSeek-V2-Lite and $B=4$ for Mixtral-8×7B. The $\Delta$ rows give the change of MaskCoFT against each reference, in percentage points for HR and as a relative change for Fetches/tok.}
  \label{tab:simulation}
  \small
  \setlength{\tabcolsep}{4pt}
  \begin{tabular}{@{}lcccccc@{}}
    \toprule
    & \multicolumn{2}{c}{LRU} & \multicolumn{2}{c}{LFU} & \multicolumn{2}{c}{FIFO} \\
    \cmidrule(lr){2-3}\cmidrule(lr){4-5}\cmidrule(lr){6-7}
    Method & HR (\%) $\uparrow$ & Fetches/tok $\downarrow$ & HR (\%) $\uparrow$ & Fetches/tok $\downarrow$ & HR (\%) $\uparrow$ & Fetches/tok $\downarrow$ \\
    \midrule
    \multicolumn{7}{@{}l}{\textit{\textbf{DeepSeek-V2-Lite  $B=12$}}} \\
    Baseline & 44.45 & 86.66 & 43.45 & 88.21 & 43.34 & 88.39 \\
    CE-only & 43.08 & 88.80 & 41.48 & 91.29 & 41.59 & 91.12 \\
    MaskCoFT (Ours) & \textbf{47.85} & \textbf{81.36} & \textbf{49.18} & \textbf{79.28} & \textbf{45.75} & \textbf{84.63} \\
    \addlinespace[2pt]
    $\Delta_{\mathrm{Baseline}}$ & $+3.40$ & $-6.1\%$ & $+5.73$ & $-10.1\%$ & $+2.41$ & $-4.3\%$ \\
    $\Delta_{\text{CE-only}}$ & $+4.77$ & $-8.4\%$ & $+7.70$ & $-13.2\%$ & $+4.16$ & $-7.1\%$ \\
    \midrule
    \multicolumn{7}{@{}l}{\textit{\textbf{Mixtral-8×7B $B=4$}}} \\
    Baseline & 61.38 & 24.72 & 61.58 & 24.59 & 60.49 & 25.29 \\
    CE-only & 60.17 & 25.49 & 60.62 & 25.21 & 59.21 & 26.10 \\
    MaskCoFT (Ours) & \textbf{67.79} & \textbf{20.61} & \textbf{70.68} & \textbf{18.77} & \textbf{65.95} & \textbf{21.79} \\
    \addlinespace[2pt]
    $\Delta_{\mathrm{Baseline}}$ & $+6.41$ & $-16.6\%$ & $+9.10$ & $-23.7\%$ & $+5.46$ & $-13.8\%$ \\
    $\Delta_{\text{CE-only}}$ & $+7.62$ & $-19.1\%$ & $+10.06$ & $-25.5\%$ & $+6.74$ & $-16.5\%$ \\
    \bottomrule
  \end{tabular}
\end{table}
Table~\ref{tab:simulation} compares the hit rate (HR) and the expert fetches per token (Fetches/tok) of both models under the LRU, LFU, and FIFO policies. Following the cache-budget setting suggested by \citet{liang2026not} to balance cache effectiveness and efficiency, we set the GPU cache size to 12 experts per layer for DeepSeek-V2-Lite and 4 experts per layer for Mixtral-8×7B. For MaskCoFT, LFU has the highest HR and the lowest Fetches/tok on both models, and the largest gains over the Baseline. Under the LFU policy, the cache removes the least frequently accessed experts from the cache.

\begin{figure}[htbp]
\centering
\includegraphics[width=\linewidth]{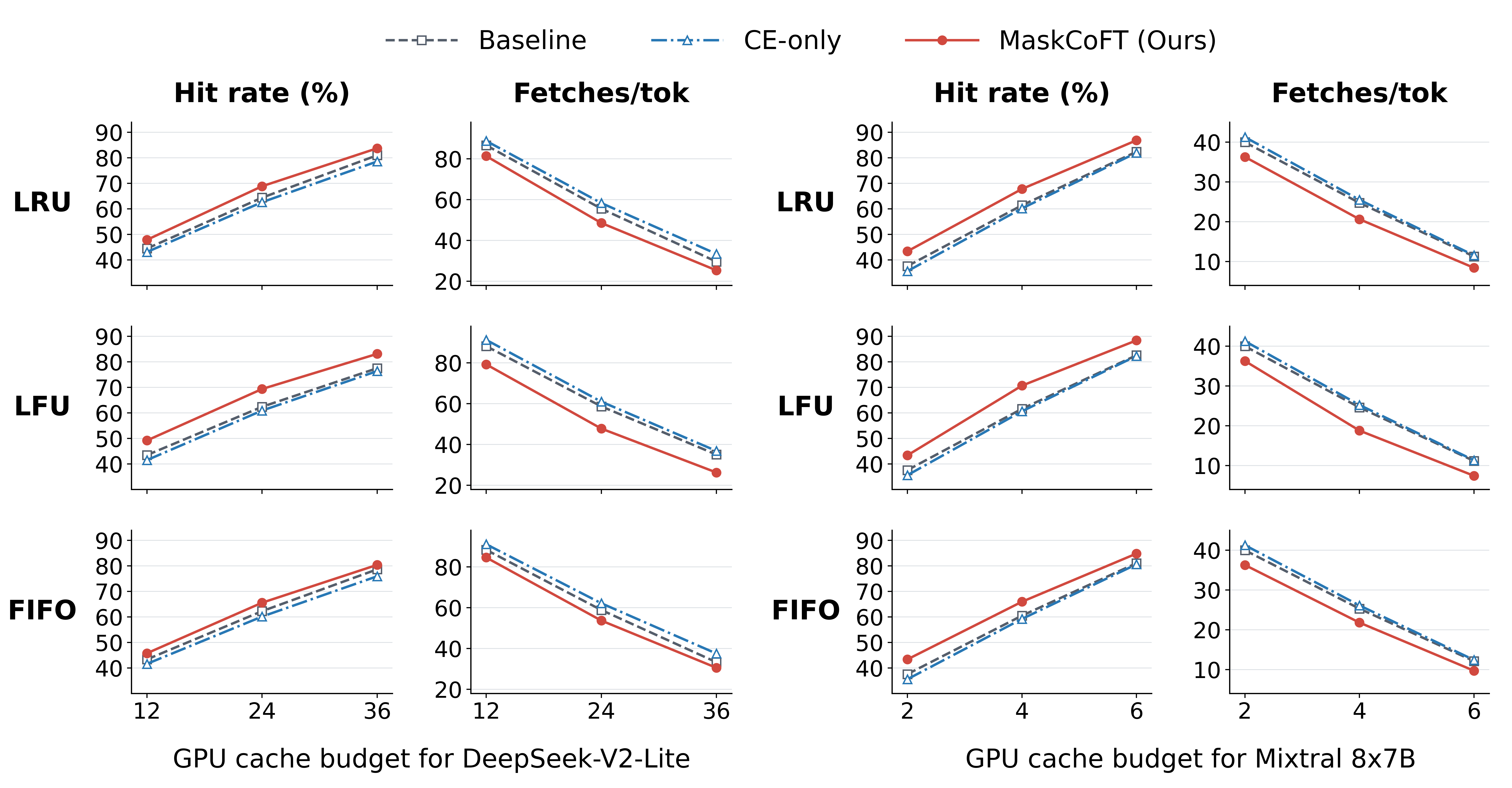}
\caption{Hit rate and Fetches/tok under LRU, LFU, and FIFO as the cache budget $B$ varies. \textbf{Left}: DeepSeek-V2-Lite with $B\in\{12,24,36\}$. \textbf{Right}: Mixtral-8x7B with $B\in\{2,4,6\}$.}
\label{fig:ablation_cache}
\end{figure}

The strong performance under LFU suggests that MaskCoFT concentrates expert activations across the decoding sequences. \textbf{In particular, the binary mask used during fine-tuning is shared across input tokens and is therefore input-independent. This encourages the router to favor a globally reusable subset of experts rather than selecting a different subset for each token.} Under LFU, the coarse-grained Mixtral-8×7B gains more HR over its base model than the fine-grained DeepSeek-V2-Lite, $+9.10$ versus $+5.73$ percentage points (Table~\ref{tab:simulation}). Under the same compact ratio of \(75\%\), this result suggests that Mixtral-8×7B can concentrate activations more strongly because its smaller routed-expert pool leaves substantially fewer candidate experts per layer after masking.

\textbf{Ablation on GPU cache sizes}
Figure~\ref{fig:ablation_cache} varies the cache budget $B$. On both models, MaskCoFT has the highest HR and the fewest Fetches/tok under all three policies at every budget. Under LFU, its relative reduction in Fetches/tok over the Baseline grows with the budget on both models.

\subsection{System Performance Measurement}
\begin{figure}[htbp]
\centering
\includegraphics[width=\linewidth]{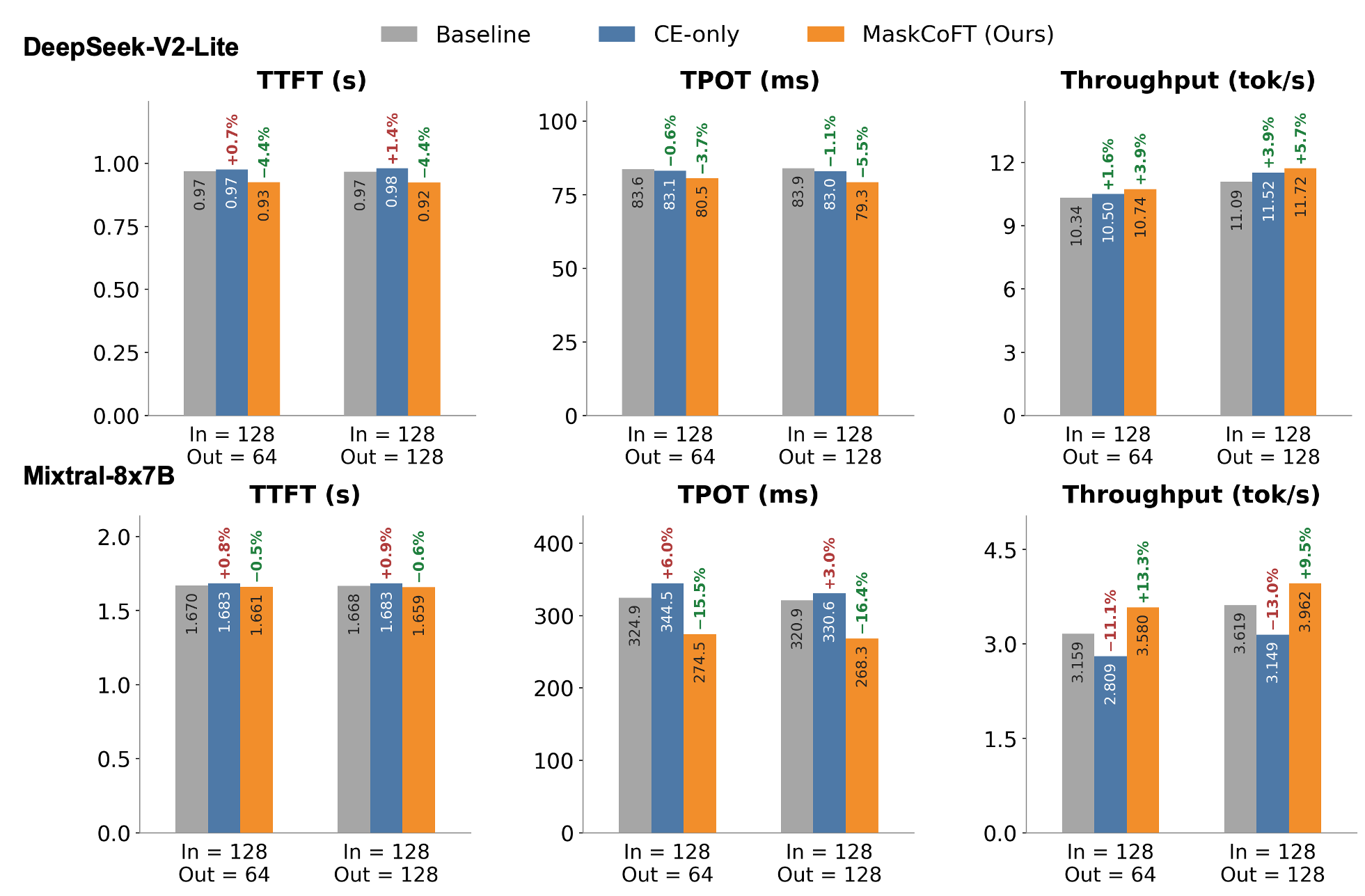}
\caption{System performance for Mixtral-8×7B (Top) DeepSeek-V2-Lite (Bottom) and  under different output token lengths.}
\label{fig:system_mixtral}
\end{figure}

Figures~\ref{fig:system_mixtral} reports the performance of end-to-end decoding in MoE-Offloading~\citep{eliseev2023fast}. On Mixtral-8×7B, MaskCoFT lowers TPOT by 15.5\% and 16.4\% for 64 and 128 output tokens and raises throughput by 13.3\% and 9.5\%. On DeepSeek-V2-Lite, it lowers TPOT by 3.7\% and 5.5\% and raises throughput by 3.9\% and 5.7\%. TTFT changes by at most 4.4\%, likely because a 128-token prefill touches most experts in each layer, regardless of the routing. The larger gain on Mixtral-8×7B is consistent with its larger cut in simulated fetches (Table~\ref{tab:simulation}). CE-only is slower than the Baseline on Mixtral-8×7B, with 3.0\% to 6.0\% higher TPOT. Refer to Appendix \ref{Appendix_D} for a decoding latency breakdown.
 
\subsection{Comparison with ReMoE}
Table~\ref{tab:compare_remoe} compares MaskCoFT with the numbers reported by ReMoE~\citep{zhu2026remoe} on DeepSeek-V2-Lite. Our fine-tuning data match those of ReMoE in source, size and preprocessing. The simulation uses the same number of ShareGPT sequences as ReMoE. ReMoE does not report its sampling seeds or the number of shots for GSM8K and MMLU. We use 5-shot GSM8K and 0-shot MMLU, and Appendix~\ref{Appendix_B} gives our sampling seed. We decode with batch size 1 to match the hit-rate metric of ReMoE. On HumanEval, whose setting ReMoE reports in full, both sources give the same Baseline and CE-only scores. The other Baseline and CE-only numbers differ, likely because of these unreported settings. We therefore compare each method with its own Baseline. ReMoE has released no checkpoint or code, so we compare only with its reported numbers.

On accuracy, ReMoE is ahead overall. The GSM8K and HumanEval differences amount to a few test problems each. Relative to its Baseline, MaskCoFT loses less on GSM8K ($-0.30$ versus $-0.68$ points) and gains less on HumanEval ($+1.22$ versus $+2.44$). The clear gap is MMLU, where ReMoE keeps its score and MaskCoFT loses 2.17 points. On hit rate, both methods gain about the same under LFU ($+5.73$ versus $+5.54$ points). Their Baseline LFU hit rates differ by 2.52 points, so we do not read this gap as a lead. ReMoE gains more under LRU ($+5.16$ versus $+3.40$) and FIFO ($+4.98$ versus $+2.41$). Its gains are similar across the three policies, while those of MaskCoFT depend on the policy. This pattern is consistent with the design of MaskCoFT. LFU keeps the experts with the most accesses, so it rewards a stable skew toward a few experts. The learned prior of MaskCoFT favors the same experts for every token. LRU and FIFO keep recently used or recently loaded experts. They reward reuse between nearby tokens, which the temporal locality regularizer of ReMoE targets directly. MaskCoFT uses no auxiliary loss. It does train the experts through LoRA and re-rank experts at inference, while ReMoE tunes only the routers.

\begin{table}[htbp]
  \centering
  \caption{Comparison with the numbers reported by ReMoE~\citep{zhu2026remoe} on DeepSeek-V2-Lite. Our runs use 5-shot GSM8K, 0-shot HumanEval pass@1 and 0-shot MMLU. GSM8K reports both the flexible-extract and the strict-match score. ReMoE does not report its GSM8K and MMLU shot counts. HR is the hit rate with a cache of $B=12$ routed experts per layer and batch size 1. The $\Delta$ rows give each method minus its own Baseline or CE-only run, in percentage points. ReMoE reports no CE-only hit rates (N/A).}
  \label{tab:compare_remoe}
  \small
  \setlength{\tabcolsep}{4pt}
  \begin{tabular}{@{}lccccccc@{}}
    \toprule
    & \multicolumn{4}{c}{Accuracy (\%)} & \multicolumn{3}{c}{HR (\%)  $B=12$} \\
    \cmidrule(lr){2-5}\cmidrule(lr){6-8}
    Method & GSM8K (flex) & GSM8K (strict) & HumanEval & MMLU & LRU & LFU & FIFO \\
    \midrule
    \multicolumn{8}{@{}l}{\textit{\textbf{Reported by ReMoE}}} \\
    Baseline & 39.04 & 38.89 & 26.83 & 57.72 & 45.19 & 45.97 & 44.32 \\
    CE-only & 37.23 & 36.92 & 28.05 & 57.44 & N/A & N/A & N/A \\
    ReMoE & 38.36 & 38.13 & 29.27 & 57.81 & 50.35 & 51.51 & 49.30 \\
    \addlinespace[2pt]
    $\Delta_{\mathrm{Baseline}}$ & $-0.68$ & $-0.76$ & $+2.44$ & $+0.09$ & $+5.16$ & $+5.54$ & $+4.98$ \\
    $\Delta_{\text{CE-only}}$ & $+1.13$ & $+1.21$ & $+1.22$ & $+0.37$ & N/A & N/A & N/A \\
    \midrule
    \multicolumn{8}{@{}l}{\textit{\textbf{Our runs}}} \\
    Baseline & 38.36 & 38.06 & 26.83 & 55.53 & 44.45 & 43.45 & 43.34 \\
    CE-only & 37.38 & 37.07 & 28.05 & 54.73 & 43.08 & 41.48 & 41.59 \\
    MaskCoFT (Ours) & 38.06 & 37.83 & 28.05 & 53.36 & 47.85 & 49.18 & 45.75 \\
    \addlinespace[2pt]
    $\Delta_{\mathrm{Baseline}}$ & $-0.30$ & $-0.23$ & $+1.22$ & $-2.17$ & $+3.40$ & $+5.73$ & $+2.41$ \\
    $\Delta_{\text{CE-only}}$ & $+0.68$ & $+0.76$ & $0.00$ & $-1.37$ & $+4.77$ & $+7.70$ & $+4.16$ \\
    \bottomrule
  \end{tabular}
\end{table}

\section{Conclusion}
We presented MaskCoFT, a masked co-adaptive fine-tuning method for MoE decoding under memory-constrained expert offloading. During fine-tuning, a learnable binary mask concentrates the routing of each layer on a subset of experts, and the experts and router co-adapt. At inference, the learned real-valued mask acts as a soft prior that re-ranks experts. 

With 4 and 12 cached experts per layer, MaskCoFT cuts expert fetches per token by up to 23.7\% on Mixtral-8×7B and 10.1\% on DeepSeek-V2-Lite. The reductions hold under three eviction policies and at every cache budget we test. In MoE-Offloading system serving, MaskCoFT lowers the time per output token by up to 16.4\% and 5.5\%, while its average accuracy stays comparable to the base model. Fine-tuning with the same cross-entropy loss without masking gains more accuracy but fetches more experts, which indicates that the savings come from the learned mask.

\subsection*{AI use statement}

We used a generative AI tool, Claude (Anthropic), to help implement experiment code, create and refine figures, and draft and polish the paper text. We also used it to check that cited works support the claims we attribute to them. We did not use generative AI tools to propose research ideas or hypotheses, design the method or experiments, formulate or prove mathematical results, generate or clean datasets, or interpret results. The authors reviewed all AI-assisted code, figures, and text and take full responsibility for the content of this paper.






\subsection*{Reproducibility statement}
Section~\ref{headings} defines MaskCoFT, including the mask sampling, the straight-through gradient and the inference rule in Eq.~(\ref{eq:inference}). Section~\ref{evaluation} describes the models, the fine-tuning data and preprocessing, the benchmarks with their shot counts and scoring, and the cache simulator with its metrics in Eq.~(\ref{eq:metrics}). Appendix~\ref{Appendix_A} lists the model specifications, and Appendix~\ref{Appendix_B} lists the training hyperparameters and the sampling seed. During the discussion period, we will share anonymized code and fine-tuned checkpoints with the reviewers and area chairs through OpenReview. We will release both publicly after publication.


\subsubsection*{Acknowledgments}
Full information on acknowledgments will be released after the paper is published.

\bibliography{iclr2027_conference}
\bibliographystyle{iclr2027_conference}

\appendix
\section{Model Details}
\label{Appendix_A}
Table \ref{tab:model_details} summarizes the specifications of the evaluated MoE models. Note that the numbers of shared, routed, and active experts are reported per layer.

\begin{table}[htbp]
  \centering
  \caption{Specifications of the MoE models. Expert counts are per MoE layer.}
  \label{tab:model_details}
  \small
  \setlength{\tabcolsep}{3pt}
  \resizebox{\textwidth}{!}{%
  \begin{tabular}{@{}lccccccccc@{}}
    \toprule
    & Dense & MoE & Shared & Routed & Active & Total & Active & Context & BF16 size \\
    Model & layers & layers & experts & experts & experts & params (B) & params (B) & length & (GB) \\
    \midrule
    DeepSeek-V2-Lite & 1 & 26 & 2 & 64 & 6 + 2 shared & 15.7 & 2.4 & 32K & $\approx$31.4 \\
    Mixtral-8x7B & 0 & 32 & 0 & 8 & 2 & 46.7 & 12.9 & 32K & $\approx$93.4 \\
    \bottomrule
  \end{tabular}}
\end{table}


\section{Hyperparameter Settings}
\label{Appendix_B}
We use seed 42 for all data sampling, including the fine-tuning and validation subsets and the simulation prompts. For both models, we set the LoRA rank to 8, alpha to 16, and LoRA dropout to 0.05. We set a linear scheduler for the experts and routers with a peak learning rate of 2e-4, and a cosine scheduler for the mask with a peak learning rate of 0.05 and a floor ratio of 0.2. For both models, we resample the binary mask at every step. We accumulate the gradient of the real-valued mask over $T$ steps and then update it once, with $T=100$ for DeepSeek-V2-Lite and $T=1$ for Mixtral-8×7B. For DeepSeek-V2-Lite, Stage~2 starts after the mask converges in Stage 1, and there are a total of 5000 training steps. For Mixtral-8×7B, Stage~2 starts after the mask converges in Stage 1, and there are a total of 3000 training steps. 


\section{Convergence of the concentration set}
\label{Appendix_C}
\begin{figure}[h]
\centering
\includegraphics[width=\linewidth]{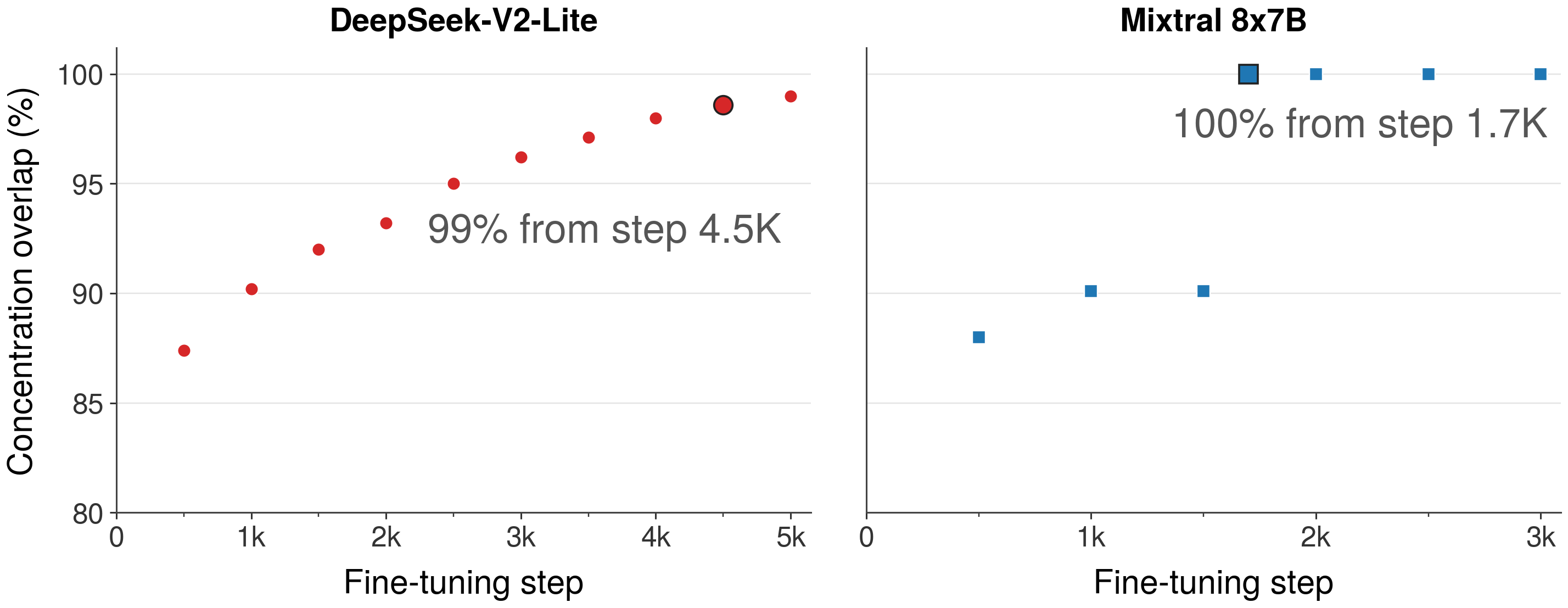}
\caption {The convergence of the concentration expert set during Stage 1 of MaskCoFT. It shows the overlap ratio between the concentration expert set and the final converged concentration set in Stage 2 every 100 steps. The mask converges early for both models. Mixtral-8×7B reaches 100$\%$ by step 1.7k, and DeepSeek-V2-Lite reaches 99$\%$ by step 4.5k.}
\label{fig:mask_convergence}
\end{figure}

\section{Decoding latency breakdown}
\label{Appendix_D}
\begin{figure}[h]
\centering
\includegraphics[width=\linewidth]{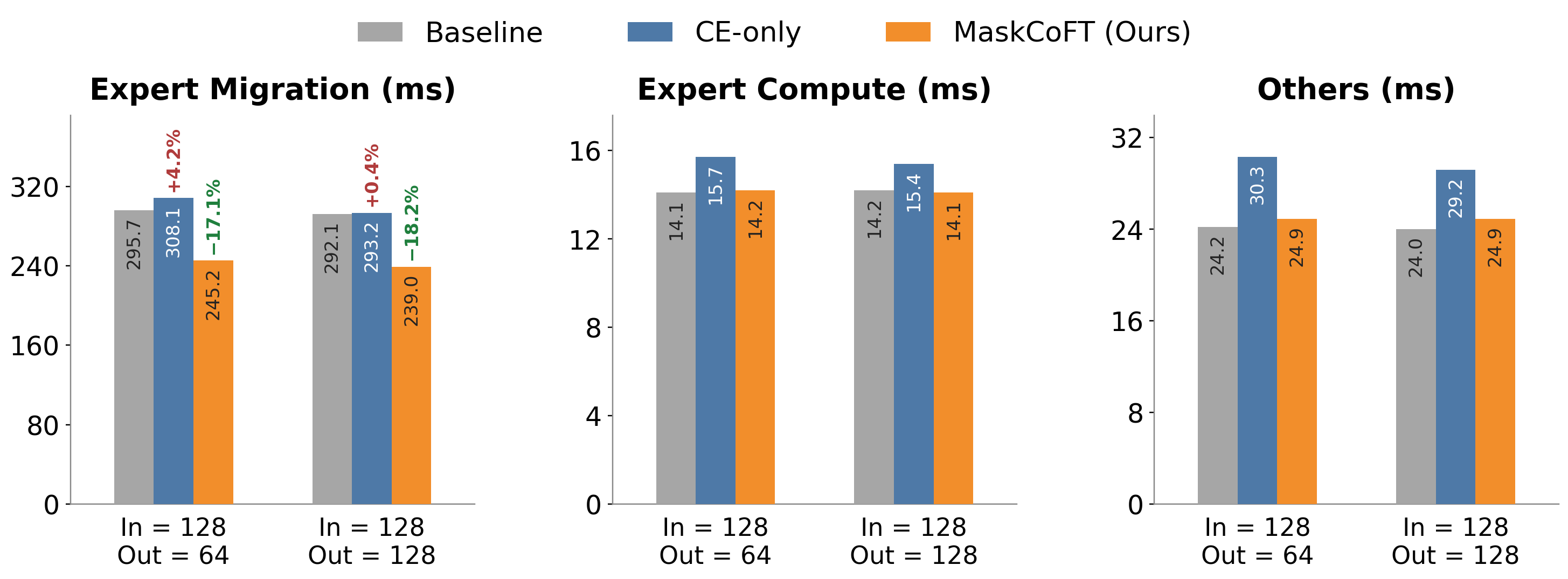}
\caption {Decoding latency breakdown for Mixtral-8×7B under different output token lengths.}
\label{fig:breakdown_mixtral}
\end{figure}

\begin{figure}[h]
\centering
\includegraphics[width=\linewidth]{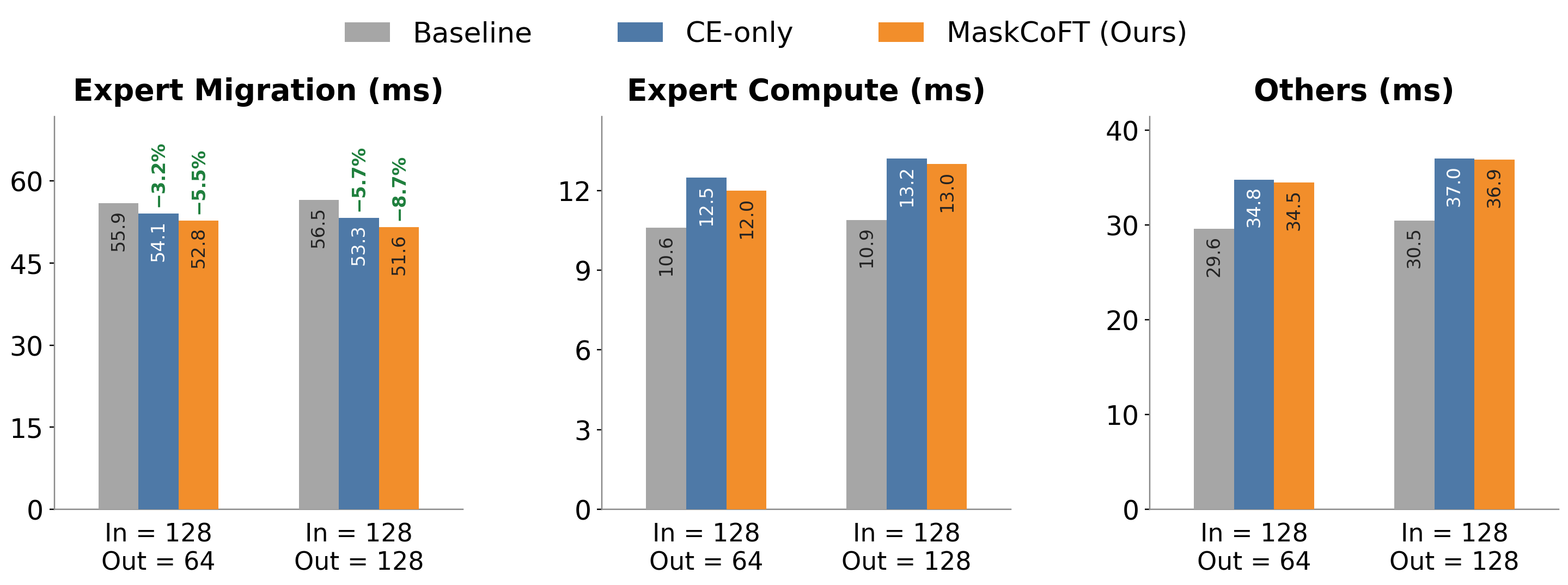}
\caption {Decoding latency breakdown for DeepSeek-V2-Lite under different output token lengths.}
\label{fig:breakdown_deepseek}
\end{figure}

\end{document}